\documentclass[nohyperref]{article}

\usepackage{microtype}
\usepackage{graphicx}
\usepackage{subcaption}
\usepackage{booktabs} 

\usepackage{hyperref}

\usepackage{dsfont}
\usepackage{bm}

\usepackage[accepted]{icml2023}

\usepackage{amsmath}
\usepackage{amssymb}
\usepackage{mathtools}
\usepackage{amsthm}

\usepackage{nicefrac}       
\usepackage{cellspace}
\usepackage{physics} 
\usepackage{multirow}

\usepackage[capitalize,noabbrev]{cleveref}

\theoremstyle{plain}
\newtheorem{theorem}{Theorem}[section]

\theoremstyle{definition}

\theoremstyle{remark}

\DeclareMathOperator*{\argmin}{arg\,min}

\usepackage[textsize=tiny]{todonotes}

\newcommand{\eb}[1]{{\scriptsize\,$\pm$\,#1}}

\vbadness=\maxdimen
\icmltitlerunning{Graph Representation Learning via Aggregation Enhancement}

\begin{document}

\twocolumn[
\icmltitle{Graph Representation Learning via Aggregation Enhancement}



\icmlsetsymbol{equal}{*}

\begin{icmlauthorlist}
\icmlauthor{Maxim Fishman}{equal,tech,int}
\icmlauthor{Chaim Baskin}{equal,tech}
\icmlauthor{Evgenii Zheltonozhskii}{equal,tech}
\icmlauthor{Almog David}{tech}
\icmlauthor{Ron Banner}{int}
\icmlauthor{Avi Mendelson}{tech}
\end{icmlauthorlist}

\icmlaffiliation{tech}{Technion}
\icmlaffiliation{int}{Habana Labs}

\icmlcorrespondingauthor{Chaim Baskin}{chaimbaskin@cs.technion.ac.il}

\icmlkeywords{Graph Neural Networks}

\vskip 0.3in
]



\printAffiliationsAndNotice{\icmlEqualContribution} 

\begin{abstract}
Graph neural networks (GNNs) have become a powerful tool for processing graph-structured data but still face challenges in effectively aggregating and propagating information between layers, which limits their performance. 
We tackle this problem with the kernel regression (KR) approach, using KR loss as the primary loss in self-supervised settings or as a regularization term in supervised settings. 
We show substantial performance improvements compared to state-of-the-art in both scenarios on multiple transductive and inductive node classification datasets, especially for deep networks. 
As opposed to mutual information (MI), KR loss is convex and easy to estimate in high-dimensional cases, even though it indirectly maximizes the MI between its inputs. 
Our work highlights the potential of KR to advance the field of graph representation learning and enhance the performance of GNNs. 
The code to reproduce our experiments is available at \url{https://github.com/Anonymous1252022/KR_for_GNNs}.
\end{abstract}
\section{Introduction}
\label{sec:intro}
\begin{figure*}
    \centering
     \includegraphics[width=\linewidth]{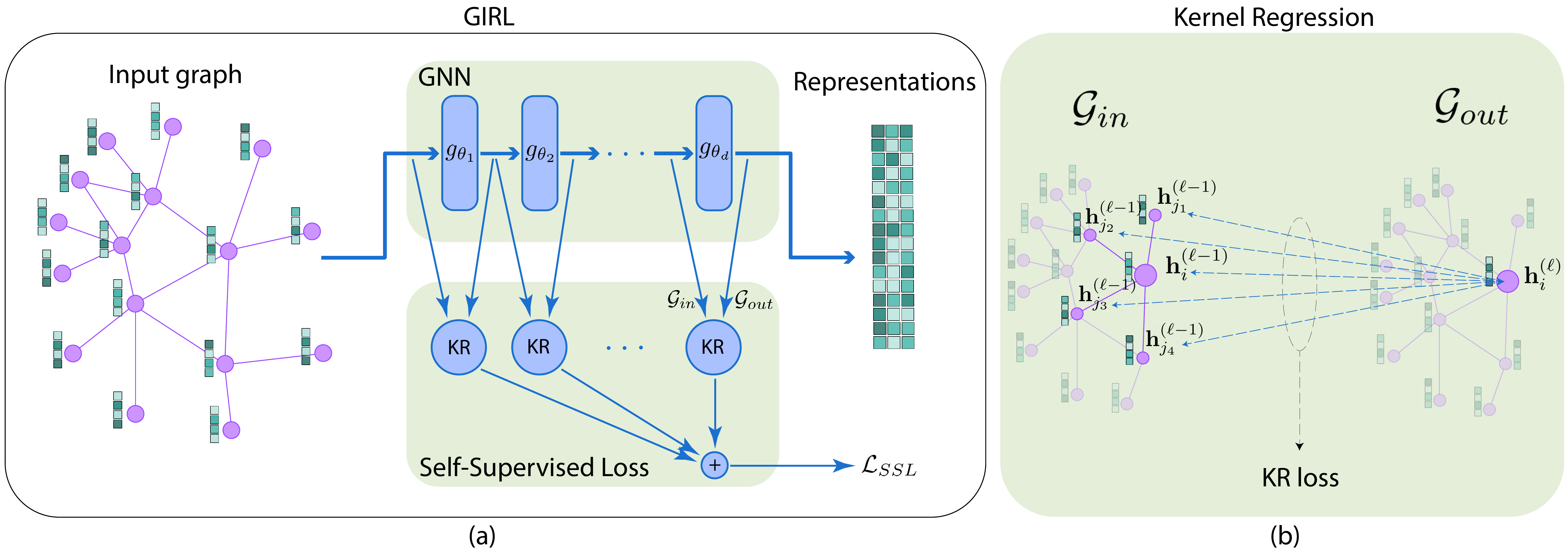}
     \vspace{-0.85cm}
    \caption{
    (a) A schematic representation of GIRL algorithm, where $\mathcal{L}_{SSL}$ is the sum of KR losses between $\mathcal{G}_{in}$ and $\mathcal{G}_{out}$ of each GNN layer $g_{\theta_\ell}$.
    (b) A schematic representation of KR. By minimizing KR loss between $\mathbf{h}_i^{(\ell)}$ (the node features of the output graph) and $\mathbf{h}_i^{(\ell - 1)}$ (node features of the input graph) as well as between $\mathbf{h}_i^{(\ell)}$ and $\qty{ \mathbf{h}_{j_k}^{(\ell - 1)} }$ (direct neighbors of $\mathbf{h}_i^{(\ell - 1)}$), we improve the information propagation and indirectly maximize MI.
    }
    \vspace{-0.3cm}
    \label{fig:Self-Supervised Setting}
\end{figure*}

\begin{figure*}
    \centering
     \includegraphics[width=\linewidth]{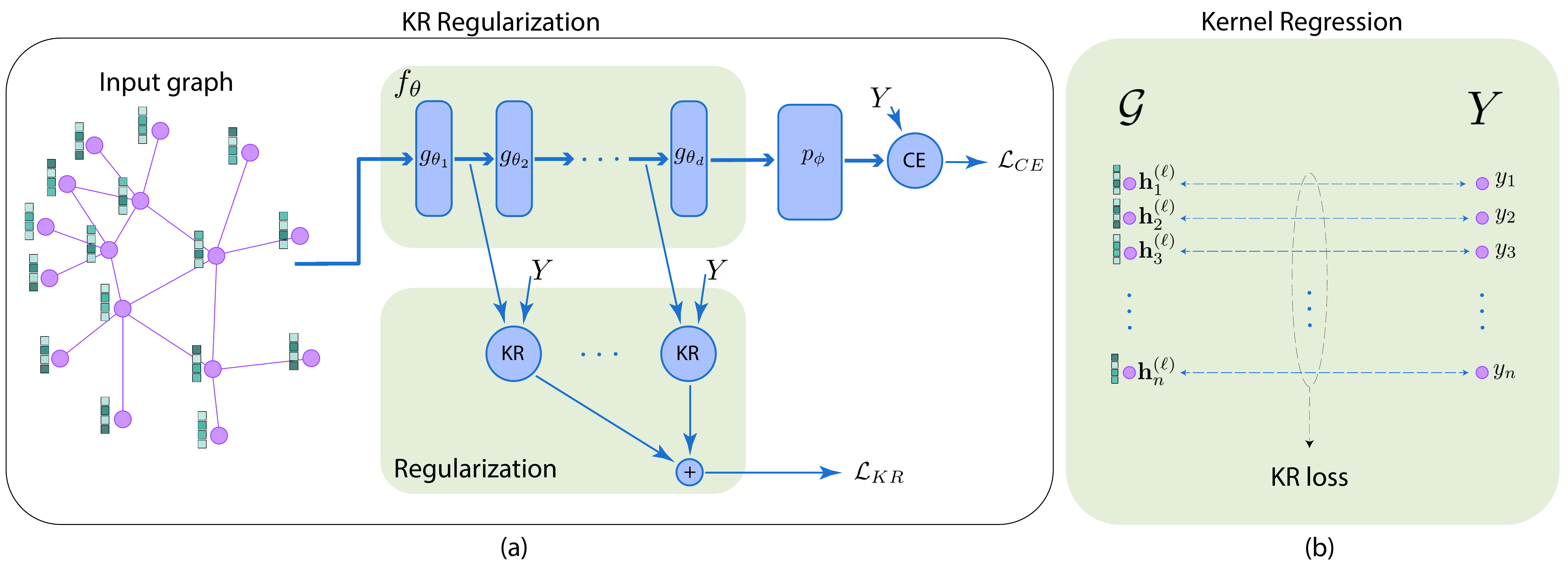}
     \vspace{-0.85cm}
    \caption{
    (a) A schematic representation of KR regularization in supervised setting, where $\mathcal{L}_{CE}$ is cross-entropy loss, and $\mathcal{L}_{KR}$ regularization loss.
    (b) A schematic representation of KR. By minimizing KR loss between $\mathbf{h}_i^{(\ell)}$ (the node features of the output graph) and $y_i$ (node classes), we improve the information propagation and indirectly maximize MI.
    }
    \vspace{-0.3cm}
    \label{fig:Supervised Setting}
\end{figure*}

Graph neural networks \citep[GNNs,][]{gori2005new,scarselli2008graph,kipf2016semi} have become a popular tool for machine learning with graphs \citep{wu2019active, ribeiro2017like, shapsoncoe2021human, cvitkovic2020supervised, sen2008collective, bloemheuvel2021computational}. 
GNNs employ a message-passing mechanism \citep{gilmer17mp},  iteratively updating node representations based on information from their neighbors to learn representations of graph-based data. 
Despite their success, GNNs still face challenges in effectively aggregating and propagating information between graph nodes. 
For example, graph convolution networks \citep[GCNs,][]{kipf2016semi} aggregate information from neighboring nodes homogeneously without the ability to selectively choose aggregation pathways.
Graph attention networks \citep[GATs,][]{velickovic2018graph} address this issue by incorporating a self-attention mechanism \citep{vaswani2017attention} into aggregation but are still prone to the \textit{depth problem} \citep{alon2021bottleneck},
the lack of ability to propagate information between distant nodes in the graph effectively.

To overcome the challenges of information aggregation, some researchers \citep{peng2020graph, bandyopadhyay2020unsupervised, sun2020infograph, velivckovic2018deep} have proposed using mutual information (MI) maximization techniques. 
MI estimation, however, can be a formidable task, especially for high-dimensional random variables \citep{10.1162/089976603321780272}, as it requires estimating high-dimensional probability density functions from a limited number of samples. 
This can be a significant limitation for deep learning approaches such as GNNs, which often involve high-dimensional representations.

In this paper, we propose utilizing the kernel regression \citep[KR,][]{https://doi.org/10.48550/arxiv.1301.5288} method as a means of enhancing graph representation learning in both supervised and self-supervised settings. 
KR is a technique that relies on reproducing kernel Hilbert space \citep[RKHS,][]{aronszajn1950theory} embedding to identify non-linear relationships between pairs of random variables.
We propose using the KR loss as surrogate for MI estimation. 
We present theoretical and empirical evidence that minimization of KR loss corresponds to the maximization of MI. 
As a result, the simplicity of KR loss estimation, coming from its convex nature, makes it a valuable alternative to direct MI maximization.
Our experimental results demonstrate that incorporating the KR loss between graph representations and the target as a regularization term (\cref{fig:Supervised Setting}) during training of deep GNNs in a supervised setting can significantly improve performance. 
Furthermore, we introduce a new self-supervised algorithm for graph representation learning called Graph Information Representation Learning (GIRL), which is based on KR and surpasses previous state-of-the-art contrast-based algorithms. 
The algorithm is shown in \cref{fig:Self-Supervised Setting}. 
We employ KR loss minimization between the input and output of each GNN layer for learning representations.

Our main contributions are as follows:
\begin{itemize}
    \item We provide evidence that KR loss minimization leads to MI maximization. 
    The convexity of KR makes it a more practical alternative to MI estimation, which can be complex, especially for high-dimensional random variables.
    
    \item We show that KR-based regularization in supervised training of deep GNNs improves information aggregation.
    
    \item We develop a self-supervised graph representation learning algorithm based on KR, named GIRL, which significantly outperforms previous state-of-the-art contrast-based algorithms.
\end{itemize}

\section{Related Work}
In this section, we briefly overview previous works on information propagation in GNNs, MI maximization, and self-supervised graph representation learning.

\paragraph{MI Maximization and KR}
MI is a measure of dependency between two random variables and is a cornerstone of information theory.
The MI maximization principle is used widely in deep learning \citep{peng2020graph, bandyopadhyay2020unsupervised, sun2020infograph}. 
MI estimation, nevertheless, remains difficult, especially for high-dimensional random variables \citep{10.1162/089976603321780272}.
\citet{kleinman2021usable} addressed this issue by introducing a notion of usable information based on MI, which is contained in the representation learned by a deep network.
In many recent works, MI is estimated via MINE \citep{belghazi2021mine} or its improvement, MI-NEE \citep{chan2019neural}, which involves a lot of additional parameters that must be learned for correct MI estimation.
Other approaches for MI estimation utilize finding lower bounds using variational Bayesian methods \citep{https://doi.org/10.48550/arxiv.1612.00410, https://doi.org/10.48550/arxiv.1711.00464,inproceedings, variational_inference}.
We eliminate explicit MI estimation in our work by turning to the KR method, a popular approach in the machine learning toolbox \citep{hofmann2008kernel, muandet2017kernel, klebanov2021conditional}.
Primarily, the KR technique has been used for finding non-linear relations between random variables by mapping their distributions into RKHS \citep{aronszajn1950theory}, where the relations turn to be linear.
Instead of finding a non-linear relation between two random variables, we use the KR loss as a proxy for MI.

\paragraph{Information Propagation to Distant Nodes}
\citet{li2018deeper} showed that the GCN model is a particular form of Laplacian smoothing and raised the concern of \textit{over-smoothing}, the inability to distinguish between node representations in deeper layers, which prevents information propagation to distant nodes.
\citet{alon2021bottleneck} proposed an alternative explanation to the problem of information propagation: GNNs are susceptible to a \textit{bottleneck} when aggregating messages across a long path. 
This bottleneck causes the \textit{over-squashing} of exponentially growing information into fixed-size vectors.
\citet{topping2021understanding} introduced a new edge-based combinatorial curvature and proved that negatively curved edges are responsible for the over-squashing issue.
\citet{https://doi.org/10.48550/arxiv.2210.00513} proposed a gradient gating framework that alleviates the over-smoothing problem.
We demonstrate that KR-based regularization leads to better information aggregation in GNNs, which improves the accuracy of deep networks.

\paragraph{Self-Supervised Learning in GNNs} 
According to \citet{Liu_2022}, graph self-supervised learning methods can be roughly split into four categories: generation-based, auxiliary property-based, contrast-based and hybrid.
The generation-based methods aim to reconstruct the input data, and can be divided into two sub-categories: feature generation that learns to reconstruct the feature information of graphs 
\cite{https://doi.org/10.48550/arxiv.2006.10141, wang_mgae:_2017, https://doi.org/10.48550/arxiv.2011.07267, https://doi.org/10.48550/arxiv.1908.02441}
and structure generation that learns to reconstruct the topological structure information of graphs \cite{https://doi.org/10.48550/arxiv.1908.07078, https://doi.org/10.48550/arxiv.1802.04407, https://doi.org/10.48550/arxiv.2204.04879, https://doi.org/10.48550/arxiv.1905.13728, https://doi.org/10.48550/arxiv.2006.02380}.
In auxiliary property-based methods, the auxiliary properties are extracted from the graph freely; afterward, the decoder aims to predict the extracted properties \cite{https://doi.org/10.48550/arxiv.2009.01674, https://doi.org/10.48550/arxiv.2003.01604, https://doi.org/10.48550/arxiv.2011.09643, multi_view, https://doi.org/10.48550/arxiv.2102.07943}.
Contrastive methods can be classified by augmentation techniques: node feature masking
\cite{zhu2020deep, https://doi.org/10.48550/arxiv.2010.13902, hu2020strategies, https://doi.org/10.48550/arxiv.2006.10141, https://doi.org/10.48550/arxiv.2105.05682}, node feature shuffling \cite{velivckovic2018deep, https://doi.org/10.48550/arxiv.1904.06316, Ren2019HDGIAU}, edge modification \cite{https://doi.org/10.48550/arxiv.2006.15437, https://doi.org/10.48550/arxiv.2006.02380, https://doi.org/10.48550/arxiv.2010.12609, https://doi.org/10.48550/arxiv.2009.05923}, graph diffusion \cite{https://doi.org/10.48550/arxiv.1911.05485, https://doi.org/10.48550/arxiv.2006.05582}, and sub-graph sampling \cite{https://doi.org/10.48550/arxiv.2009.10273}. 
Finally, hybrid methods  \cite{https://doi.org/10.48550/arxiv.2102.12380, https://doi.org/10.48550/arxiv.2009.07111, https://doi.org/10.48550/arxiv.2105.06715, https://doi.org/10.48550/arxiv.2106.04113, Jing_2021, https://doi.org/10.48550/arxiv.2111.10698, https://doi.org/10.48550/arxiv.2104.13014, https://doi.org/10.48550/arxiv.2102.06514} combine one or more of the aforementioned methods.

The most common self-supervised methods are generally contrastive, based on MI maximization and data augmentation techniques.
We propose GIRL, a novel self-supervised learning algorithm that does not require augmentations or a decoder for MI estimation.
\section{Background}
\label{sec:background}
\begin{figure}
    \centering
     \includegraphics[width=\linewidth]{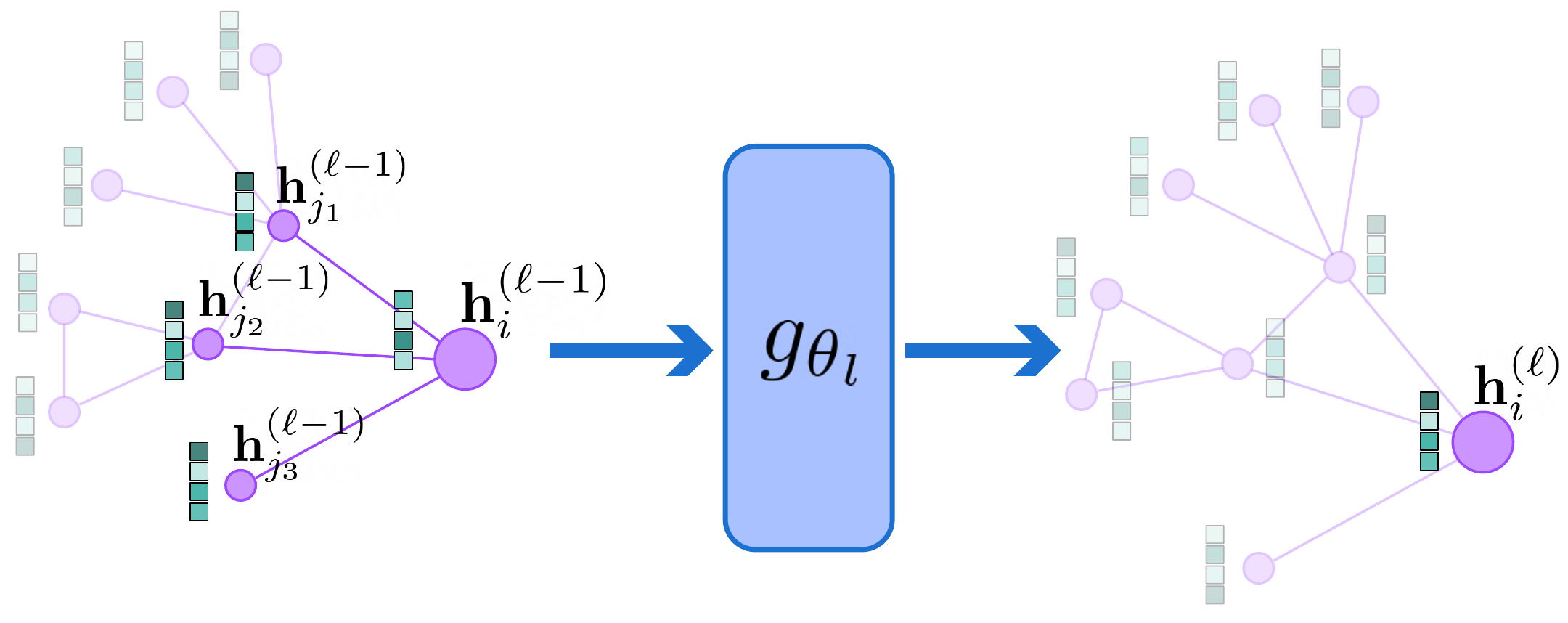}
    \caption{
    A single intermediate GNN layer $g_{\theta_l}$ receives node features $\mathbf{h}_i^{(l-1)}$ with direct neighboring node features 
    $\{\mathbf{h}_{j_1}^{(l-1)}, \mathbf{h}_{j_2}^{(l-1)}, \mathbf{h}_{j_3}^{(l-1)}\}$ as input and outputs node embedding $\mathbf{h}_i^{(l)}$.
    }
    \vspace{-0.3cm}
    \label{fig:single_layer}
\end{figure}

This section introduces the notation used throughout the paper and defines a probabilistic model for a single intermediate GNN layer. We also define the random variables involved in the information aggregation process.

\subsection{Basic Notations}
\label{sec:basic_notations}
This paper focuses on node-attributed graphs and node classification tasks for evaluating both supervised (as a primary task) and self-supervised learning (as a downstream task) performance. We use the notation $[n] = \qty{ 1, 2, \dots , n }$ throughout the paper.

A node-attributed graph is a tuple $\mathcal{G} = (\qty{\mathbf{x}_i}_{i \in [n]}, \mathcal{E})$, where $\qty{\mathbf{x}_i}_{i \in [n]} \subset \mathbb{R}^{d_1}$ are the node features and $\mathcal{E}\subseteq [n]\times [n]$ ($|\mathcal{E}|=m$) is the set of edges. The set of direct neighboring nodes for node $i$ is denoted as $\mathcal{N}(i)$.

The node classification task involves predicting the label $y_i \in \mathcal{Y}$ for each node $i \in [n]$, where $\mathcal{Y}$ is the set of possible labels. The graph encoder $f_\theta$, parameterized by $\theta$, processes the node-attributed graph $\mathcal{G}$ and generates node representations (embeddings) $\qty{\mathbf{h}_i}_{i \in [n]} \subset \mathbb{R}^{d_2}$: 
\begin{align}
    f_{\theta} (\mathcal{G}) = (\{\mathbf{h}_i\}_{i \in [n]}, \mathcal{E}).
\end{align}
 The classifier $p_{\phi}$, parameterized by $\phi$, takes these embeddings as inputs and produces logits. Finally, the cross-entropy (CE) loss $\mathcal{L}_{\text{CE}}$ is optimized. 
The node classification problem is defined as: 
\begin{align}
    \theta^*, \phi^* = \argmin_{\theta, \phi} \frac{1}{n}\sum_{j \in [n]}\mathcal{L}_{\text{CE}}(p_\phi([f_\theta(\mathcal{G})]_j), y_j),
\end{align}
where $[\cdot]_j$ is a choice function defined as $[(\{\mathbf{h}_i\}_{i \in [n]}, \mathcal{E})]_j = \mathbf{h}_j$.

The graph encoder $f_\theta$ is typically composed of multiple GNN layers, represented as: 
\begin{align} 
f_\theta = g_{\theta_{d}} \circ \cdots \circ g_{\theta_{\ell}} \circ \cdots \circ g_{\theta_{1}}, 
\end{align} 
where $d$ is the depth of the network, $g_{\theta_{\ell}}$ is an intermediate GNN layer: 
\begin{align}
    g_{\theta_{\ell}} (\{\mathbf{h}^{(\ell - 1)}_i\}_{i \in [n]}, \mathcal{E}) = (\{\mathbf{h}^{(\ell)}_i\}_{i \in [n]}, \mathcal{E}),
\end{align}
and $\theta = \qty{\theta_{1}, \dots, \theta_{\ell}, \dots, \theta_{d}}$ are the network parameters. Each GNN layer $g_{\theta_{\ell}}$ follows a message passing scheme \citep{gilmer17mp}, where information is only aggregated from direct neighboring nodes, as visualized in \cref{fig:single_layer}. As a result, the depth of the aggregation is equal to $d$, the number of layers in the graph encoder $f_\theta$.

In the self-supervised setting, the goal is to learn a meaningful graph representation without access to the node labels $\{ y_i \}_{i \in [n]}$. To achieve this, a self-supervised algorithm is used to learn the parameters $\theta^*$ of graph encoder $f_{\theta}$. The quality of the learned representations $\{\mathbf{h}_i\}_{i \in [n]}$ is then evaluated by training a network for the downstream task (in our case, node classification), with the (frozen) extracted features as input and node labels as outputs: 
\begin{align}
    \phi^* = \argmin_{\phi} \frac{1}{n}\sum_{i \in [n]}\mathcal{L}_{\text{CE}}(p_\phi([f_{\theta^*}(\mathcal{G})]_i), y_i).
\end{align}

\subsection{Probabilistic Model of Information Aggregation}
\label{sec:probabilistic_model}
We now formalize the information aggregation and propagation processes in GNNs using random variables. Given an intermediate graph representation, i.e., the output of layer $g_{\theta_\ell}$ as shown \cref{fig:single_layer}, $(\{\mathbf{h}^{(\ell)}_i\}_{i \in [n]}, \mathcal{E})$ and the node labels ${ y_i }$, we define the random variables for the node features, node neighborhood features, and node labels: $H^{(\ell)}$, $Z^{(\ell)}$, and $Y$, correspondingly. For $j$ uniformly distributed over $[n]$, $H^{(\ell)}=\mathbf{h}^{(\ell)}_j$, $Z^{(\ell)}$ is a vector uniformly distributed over $\{\mathbf{h}^{(\ell)}_k \}_{k \in \mathcal{N}(j)}$, and $Y=y_j$. The three variables are correlated, i.e., they share the same sampled $j$.

At each layer $g_{\theta_\ell}$, information flows from both $H^{(\ell-1)}$ and $Z^{(\ell-1)}$ to $H^{(\ell)}$. The random variable $Z^{(\ell-1)}$ contains the information from the neighborhood and is vital for the aggregation process.
\section{Aggregation Enhancement}
\label{subsec:recover}
Based on definitions provided in \cref{sec:background}, we present MI- and KR-based approaches for information aggregation in supervised and self-supervised settings.

\subsection{MI Approach}
The ability of the classifier $p_{\phi}$ to learn from the node representations $\{ \mathbf{h}_i \}_{i \in [n]}$ depends on the mutual information $I(H^{(d)}; Y)$ (the random variable of node representations $H^{(d)}$ was defined in \cref{sec:probabilistic_model}).
When $I(H^{(d)}; Y)=0$, the random variables $H^{(d)}$ and $Y$ are statistically independent, and it impossible to learn $Y$ from $H^{(d)}$. On the other hand, when a deterministic continuous map from $H^{(d)}$ to $Y$ exists, the random variables $H^{(d)}$ and $Y$ are dependent, resulting in a high value of the mutual information $I(H^{(d)}; Y)$. 
As a result, maximization of the total mutual information $\sum_{\ell \in [d]} I(H^{(\ell)}; Y)$ along with the supervised loss minimization enhances information aggregation. 

In self-supervised representation learning, it is unknown which information should be preserved in the node embeddings to perform the downstream task well (node classification). 
In this case, we aim to retain as much information as possible in the intermediate node representations. This can be done by maximization of the following quantity:
\begin{align}
    \sum_{l \in [d]} \qty(I(H^{(\ell)}; H^{(\ell-1)}) + I(H^{(\ell)}; Z^{(\ell-1)}))
\end{align} 

Estimating mutual information from samples, however, can be difficult, especially for high-dimensional random vectors. We propose using kernel regression \citep[KR,][]{https://doi.org/10.48550/arxiv.1301.5288} as an alternative, which is easier to estimate and does not require additional learnable parameters.

\subsection{The KR Approach}
In the following, we redefine KR loss $\rho$ and convince the reader that $\rho$ minimization leads to MI maximization. 
Next, we demonstrate the KR loss minimization technique for aggregation enhancement in GNNs.

As previously noted, the ability to learn node labels $\{ y_i \}_{i \in [n]}$ from node representations $\{\mathbf{h}^{(d)}_i\}_{i \in [n]}$ depends on the existence of a continuous map from $H^{(d)}$ to $Y$. 
Let $X$ and $Y$ be two random variables with values in $\mathbb{R}^\ell$ and $\mathbb{R}^m$, respectively. 
We define
\begin{align}
    \label{eq:C_X}
    \mathcal{C}_X = \qty{ f(X) \;|\; f:\mathbb{R}^{\ell} \rightarrow \mathbb{R}^m \text{ is continuous function}  },
\end{align}
 and evaluate the distance between the random variable $Y$ and the set $\mathcal{C}_X$:
\begin{align}
    \rho(Y|X) = \inf_{Z \in \mathcal{C}_X} d(Y, Z),
\end{align}
where $d$ is some distance between two random variables, e.g., a metric induced by the $L_p$ norm, $\mathbb{E}[\abs{X}^p]^{\nicefrac{1}{p}}$.

\begin{theorem} \label{thm:t1} 
For any pair of random variables $X$ and $Y$, $\rho(Y|X)=0$ if and only if $I(X;Y)=H(Y)$,
where $I(X;Y)$ is the MI between $X$ and $Y$, and $H(Y)$ is the entropy of $Y$. 
\end{theorem} 
The proof of \cref{thm:t1} is given in \cref{proof:t1}.

The consequence of \cref{thm:t1} is that for a fixed $Y$, finding random variable $X$ such that $\rho(Y|X)=0$ is equivalent to finding an $X$ that maximizes the $I(X;Y)$, 
because $I(X;Y) \leq H(Y)$ for any two random variables $X$ and $Y$.
Additionally, the empirical result given in \cref{fig:KL_vs_MI} presents evidence that KR loss minimization leads to MI maximization.

The following theorem shows how KR loss $\rho$ can be efficiently estimated.
\begin{theorem} 
    \label{thm:t2}
    Given a collection of points $\{ \mathbf{x}_i \}_{i \in [n]}$ and $\{ \mathbf{y}_i \}_{i \in [n]}$ sampled i.i.d. from $X$ and $Y$, respectively, the empirical estimation of distance $\rho(Y|X)$ is given by
    \begin{align}
        \hat{\rho}(Y|X) 
        = 
        \frac{1}{m}\sum_{i \in [m]} \left( 
            \left(
                \frac{1}{n}\sum_{k \in [n]} 
                \left|
                    \sum_{j \in [n]} (\mathbb{I} - \Pi)_{kj}\mathbf{y}_j
                \right|^p
            \right)^{\nicefrac{1}{p}}
        \right)_i
    \end{align}
    where $\Pi$ is an orthogonal projection onto $\Im(K)$
    and $K_{ij} = \exp\left( -\frac{\norm{\mathbf{x}_i - \mathbf{x}_j}^2}{2 \sigma^2} \right)$ is a Gram matrix.
\end{theorem}
We provide the proof of \cref{thm:t2} in \cref{proof:t2} and experiments on synthetic data in \cref{exp:synthetic_data}.

Turning back to the graph supervised setting, we propose to replace maximization of total mutual information $\sum_{l \in [d]} I(H^{(l)}; Y)$ by minimization of $\sum_{l \in [d]} \hat{\rho}(Y|H^{(l)})$, as follows:
\begin{align}
    \label{eq:kr_supervised}
    \theta^*, \phi^* = \argmin_{\theta, \phi} \frac{1}{n}\sum_{j \in [n]}\mathcal{L}_{\text{CE}}(p_\phi([f_\theta(\mathcal{G})]_j), y_j) + \\ 
    +\lambda\sum_{l \in [d]} \hat{\rho}(Y|H^{(l)})
\end{align}

We also propose a self-supervised graph representation learning algorithm called GIRL (Graph Information Representation Learning). In the self-supervised setting, our goal is to transfer as much information as possible from the random variables $H^{(\ell-1)}$ and $Z^{(\ell-1)}$ to $H^{(l)}$ in each layer $\ell \in [d]$ of the graph encoder $f_{\theta}$. To do this, we seek to minimize the total distance $\rho(H^{(l-1)}|H^{(l)}) + \rho(Z^{(l-1)}|H^{(l)})$ between the current layer node embeddings and previous layer embeddings of the node and its neighbors. This is achieved by solving the following optimization problem: 
\begin{align}
    \theta^* = \argmin_{\theta} \sum_{l \in [d]} 
    \left(
        \hat{\rho}(H^{(l-1)}|H^{(l)}) + \hat{\rho}(Z^{(l-1)}|H^{(l)})
    \right)
\end{align}
The algorithm is provided in \cref{alg:GIRL} and illustrated in \cref{fig:Self-Supervised Setting}.

\section{Experiments}

\begin{figure}
    \centering
     \includegraphics[width=\linewidth]{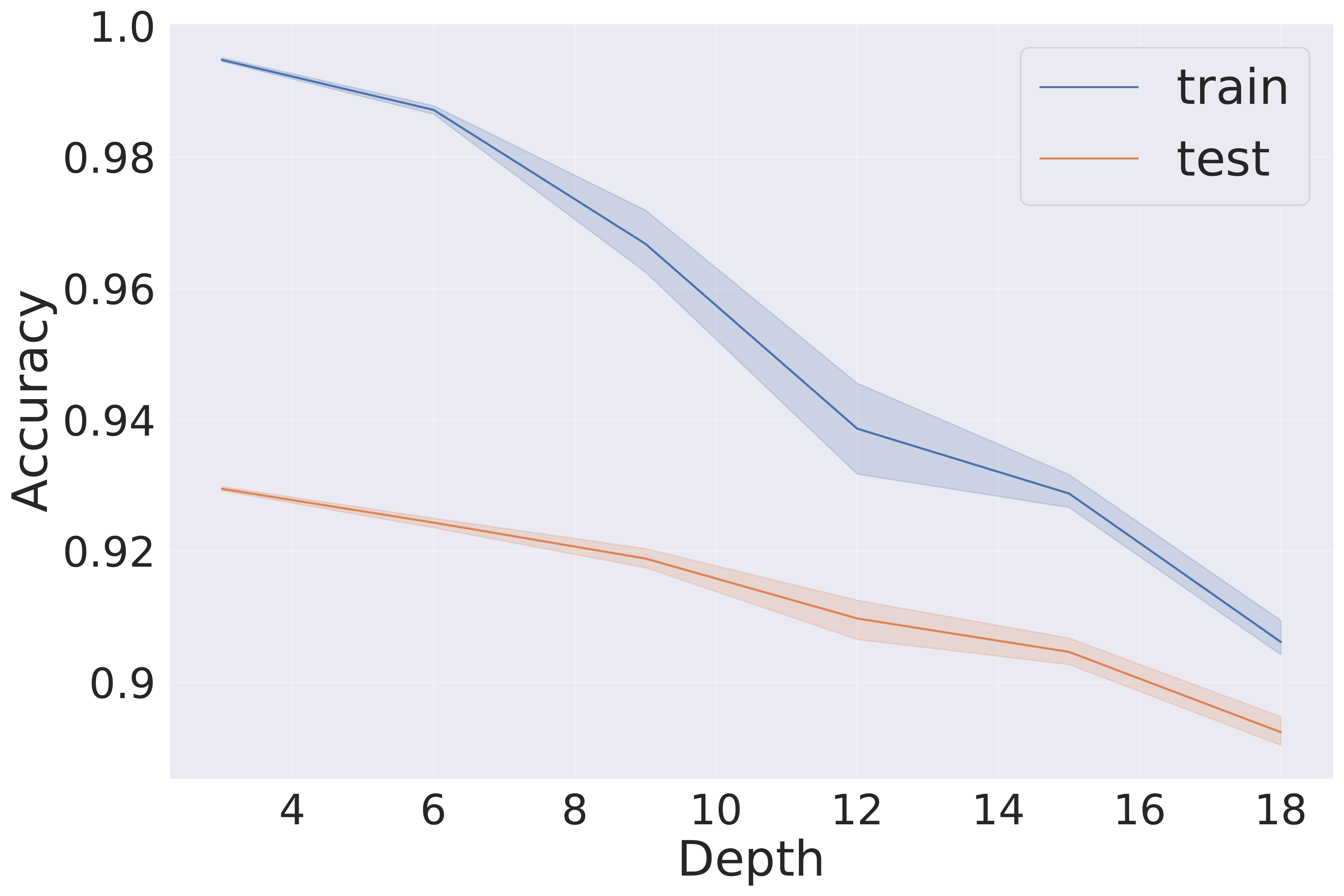}
     \vspace{-0.85cm}
    \caption{Reddit: The drop in training and test accuracy, along with the GNN depth extension.}
     \vspace{-0.3cm}
    \label{fig:depth_problem1}
\end{figure}
\begin{figure}
    \centering
     \includegraphics[width=\linewidth]{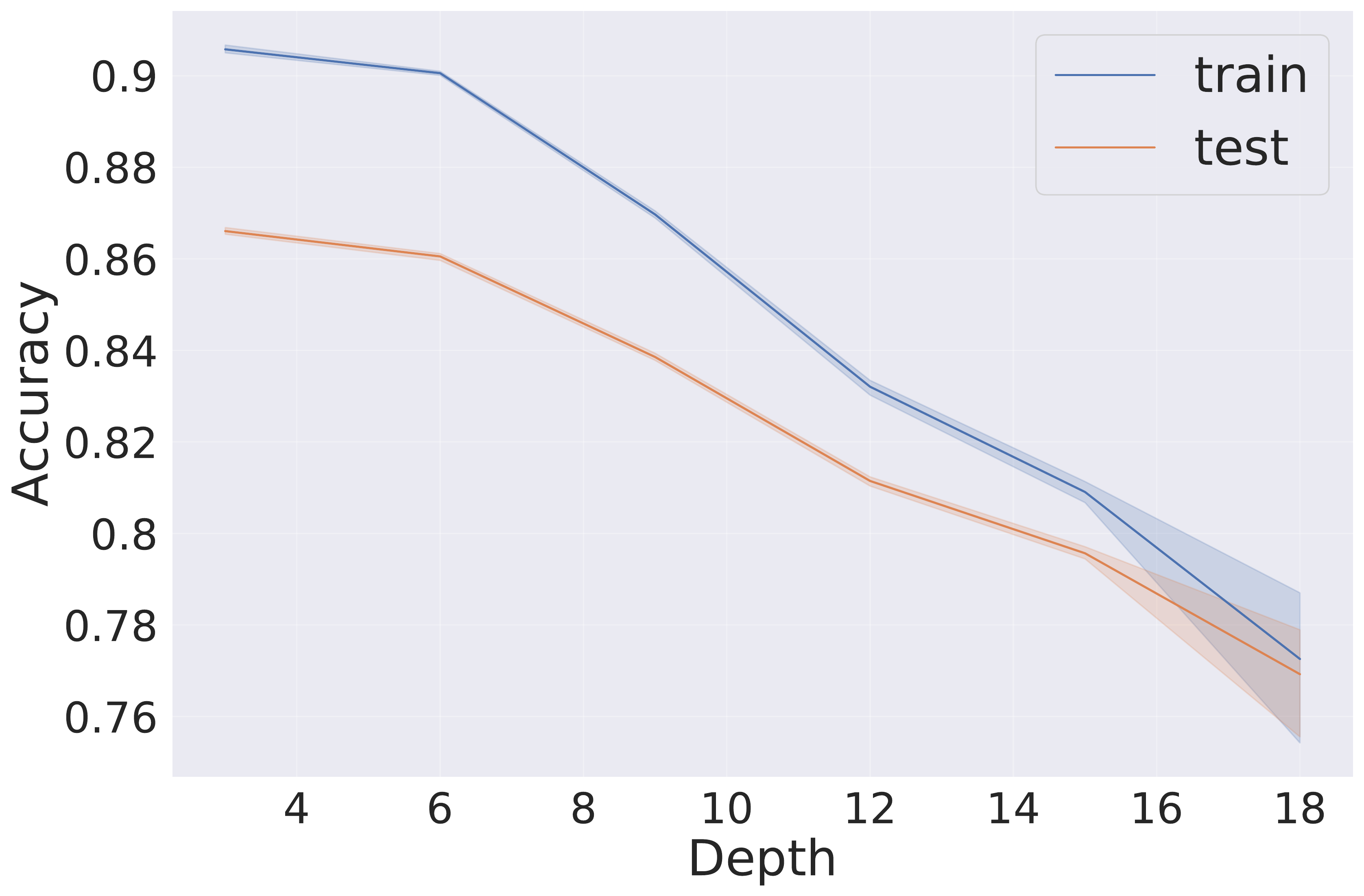}
     \vspace{-0.85cm}
    \caption{PPI: The drop in training and test accuracy, along with the GNN depth extension.}
     \vspace{-0.3cm}
    \label{fig:depth_problem2}
\end{figure}

\begin{table*}
    \centering
    \begin{tabular}{ccccccccc} 
        \toprule
        \multirow{2}{*}{\textbf{Depth}}     & \multirow{2}{*}{\textbf{Method}}  & \multicolumn{7}{c}{\textbf{Accuracy}}           \\ 
        
        \cline{3-9} 
                                            &           & \textbf{Reddit}       & \textbf{Reddit2}      & \textbf{ogbn-arxiv}   & \textbf{ogbn-products}& \textbf{PPI}          & \textbf{Texas}        & \textbf{Actor}    \\
                                            
        \specialrule{0.5pt}{0.5pt}{1pt}
	    \midrule
        \multirow{4}{*}{3}                  & None      & 92.9\eb{0.1}          & 93.0\eb{0.1}          & \textbf{66.2\eb{0.3}} & 75.5\eb{0.3}          & 86.6\eb{0.1}          & \textbf{77.8\eb{1.2}} & 34.2\eb{0.8}      \\
                                            & WD        & 92.9\eb{0.1}          & 92.9\eb{0.1}          & 65.9\eb{0.3}          & 75.6\eb{0.3}          & 85.9\eb{0.1}          & 76.2\eb{2.3}          & 34.4\eb{0.6}      \\
                                            & +FA       & 92.6\eb{0.1}          & 92.0\eb{0.1}          & 65.8\eb{0.4}          & 74.3\eb{0.1}          & 85.8\eb{0.1}          & 73.0\eb{3.3}          & 33.6\eb{1.3}      \\
                                            & KR (ours) & \textbf{93.0\eb{0.1}} & \textbf{93.1\eb{0.1}} & \textbf{66.2\eb{0.4}} & \textbf{75.6\eb{0.2}} & \textbf{86.8\eb{0.1}} & 77.3\eb{2.3}          & \textbf{35.1\eb{0.3}}\\
        \cline{1-9}
        \multirow{4}{*}{9}                  & None      & 91.9\eb{0.2}          & 90.3\eb{0.6}          & 66.4\eb{0.5}          & 75.8\eb{0.3}          & 83.9\eb{0.1}          & 62.7\eb{7.0}          & 31.8\eb{3.9}      \\
                                            & WD        & 91.5\eb{0.3}          & 89.7\eb{1.1}          & 65.3\eb{0.4}          & 75.3\eb{0.5}          & 82.2\eb{0.2}          & 68.1\eb{3.0}          & 31.3\eb{1.5}      \\
                                            & +FA       & \textbf{92.8\eb{0.1}} & 90.6\eb{0.6}          & 66.9\eb{0.1}          & \textbf{76.3\eb{0.1}} & 83.7\eb{0.1}          & 37.8\eb{11.0}         & 32.7\eb{0.9}      \\
                                            & KR (ours) & 92.6\eb{0.1}          & \textbf{92.1\eb{0.1}} & \textbf{67.2\eb{0.3}} & 75.8\eb{0.1}          & \textbf{85.0\eb{0.1}} & \textbf{75.7\eb{2.7}} & \textbf{32.8\eb{0.4}}\\
        \cline{1-9}
        \multirow{4}{*}{18}                 & None      & 89.2\eb{0.3}          & 33.3\eb{25.7}         & 64.7\eb{0.2}          & 73.2\eb{0.3}          & 76.9\eb{1.5}          & 63.8\eb{1.5}          & 23.8\eb{2.1}      \\
                                            & WD        & 88.4\eb{0.2}          & 14.9\eb{0.0}          & 63.4\eb{0.2}          & 72.4\eb{0.9}          & 74.3\eb{0.1}          & 61.1\eb{8.5}          & 24.9\eb{0.9}          \\
                                            & +FA       & 82.5\eb{1.7}          & 52.6\eb{21.7}         & 57.2\eb{1.9}          & 63.5\eb{7.1}          & 78.7\eb{0.1}          & 62.2\eb{12.1}         & 24.6\eb{1.1}      \\
                                            & KR (ours) & \textbf{91.4\eb{0.4}} & \textbf{76.4\eb{1.0}} & \textbf{65.7\eb{0.3}} & \textbf{74.6\eb{0.4}} & \textbf{80.2\eb{0.2}} & \textbf{67.6\eb{2.4}} & \textbf{30.2\eb{1.1}}\\
        \bottomrule
    \end{tabular}

    \caption{
    The performance of the model with different depths, both with and without weight decay (WD), fully-adjacent layer (+FA) \citep{alon2021bottleneck},
    and kernel regression loss (KR). 
    The table is divided into three blocks based on depth, with the best results for each depth highlighted in bold.
    }
    \vspace{-0.3cm}
    \label{tab:depth_problem}
\end{table*}

\begin{table*}[t]
    \centering
    \begin{tabular}{ccccccc}
    \multicolumn{7}{c}{(a) \textit{Transductive}} \\
    \toprule
    \textbf{Dataset}    & \textbf{DGI}  & \textbf{GRACE}& \textbf{BGCL} & \textbf{BGRL} & \textbf{GraphMAE} & \textbf{GIRL (ours)}  \\
    \specialrule{0.5pt}{0.5pt}{1pt}
    \midrule
    Cora                & 82.3\eb{0.6}  & 84.0\eb{0.1}  & 83.8\eb{0.3}  & -             & 84.2\eb{0.4}      & \textbf{88.3\eb{0.1}} \\
    Citeseer            & 71.8\eb{0.7}  & 72.1\eb{0.5}  & 72.7\eb{0.3}  & -             & 73.4\eb{0.4}      & \textbf{79.1\eb{0.3}} \\
    Pubmed              & 76.8\eb{0.6}  & 86.7\eb{0.1}  & -             & -             & 81.1\eb{0.4}      & \textbf{89.0\eb{0.0}} \\
    DBLP                & -             & 84.2\eb{0.1}  & -             & -             & -                 & \textbf{85.9\eb{0.1}} \\
    Amazon-Photos       & 91.6\eb{0.2}  & 92.2\eb{0.2}  & 92.5\eb{0.2}  & 93.2\eb{0.3}  & -                 & \textbf{95.6\eb{0.1}} \\
    WikiCS              & 75.4\eb{0.1}  & 80.1\eb{0.5}  & -             & 80.0\eb{0.1}  & -                 & \textbf{83.3}\eb{0.0} \\
    Amazon-Computers    & 84.0\eb{0.5}  & 89.5\eb{0.4}  & -             & 90.3\eb{0.2}  & -                 & \textbf{91.8}\eb{0.0} \\
    Coauthor CS         & 92.2\eb{0.6}  & 91.1\eb{0.2}  & -             & 93.3\eb{0.1}  & -                 & \textbf{94.5}\eb{0.0} \\
    Coauthor Physics    & 94.5\eb{0.5}  & -             & -             & 95.7\eb{0.1}  & -                 & \textbf{96.6}\eb{0.0} \\
    \bottomrule
    \end{tabular}
    \\
    \begin{tabular}{ccccccccc}
    \multicolumn{7}{c}{} \\
	\multicolumn{7}{c}{(b) \textit{Inductive}} \\
    \toprule
    \textbf{Dataset}& \textbf{DGI}  & \textbf{GRACE}        & \textbf{BGRL}         & \textbf{SimGRACE}     & \textbf{GraphMAE}     & \textbf{GIRL (ours)}  \\
    \specialrule{0.5pt}{0.5pt}{1pt}
    \midrule
    PPI             & 63.8\eb{0.2}  & 66.2\eb{0.1}          & 70.5\eb{0.1}          & 70.3\eb{1.2}          & 74.5\eb{0.3}          & \textbf{87.9\eb{0.1}} \\
    Reddit          & 94.0\eb{0.1}  & \textbf{94.2\eb{0.0}} & -                     & -                     & -                     & 92.1\eb{0.0}          \\
    Reddit2         & -             & 91.7\eb{0.0}$^*$      & -                     & -                     & \textbf{96.0\eb{0.1}} & 92.1\eb{0.0} \\
    ogbn-arxiv      & -             & 69.0\eb{0.3}$^*$      & 71.6\eb{0.1}          & -                     & 71.75\eb{0.17}        & \textbf{71.8\eb{0.2}} \\
    ogbn-products   & -             & 68.1\eb{0.1}$^*$      & -                     & -                     & -                     & \textbf{71.5\eb{0.3}} \\
    \bottomrule
    \end{tabular}
    \\
    \begin{tabular}{ccccccccc}
    \multicolumn{4}{c}{} \\
	\multicolumn{4}{c}{(c) \textit{Heterophilic}} \\
    \toprule
    \textbf{Dataset}& \textbf{DGI}  & \textbf{SELENE}     & \textbf{GIRL (ours)}  \\
    \specialrule{0.5pt}{0.5pt}{1pt}
    \midrule
    Texas           & 54.1          & 64.3          & \textbf{75.7} \\
    Actor           & 27.2          & 34.1 & \textbf{34.2} \\
    USA-Airports    & 31.1          & 56.5          & \textbf{65.1} \\
    \bottomrule
    \end{tabular}
    \\
    \begin{tabular}{cccccccc}
    \multicolumn{8}{c}{} \\
	\multicolumn{8}{c}{(d) \textit{Graph Property Prediction}} \\
    \toprule
    \textbf{Dataset}    & \textbf{InfoGraph}    & \textbf{GraphCL}  & \textbf{JOAO} & \textbf{JOAOv2}   & \textbf{SimGRACE} & \textbf{GraphMAE}     & \textbf{GIRL (ours)}  \\
    \specialrule{0.5pt}{0.5pt}{1pt}
    \midrule
    NCI1                & 77.9\eb{1.1}          & 77.9\eb{0.4}      & 78.1\eb{0.5}  & 78.4\eb{0.5}      & 79.1\eb{0.4}      & 80.4\eb{0.3} & \textbf{80.6\eb{0.1}} \\
    PROTEINS            & 74.4\eb{0.3}          & 74.4\eb{0.5}      & 74.6\eb{0.4}  & 74.1\eb{1.1}      & 75.4\eb{0.1}      & 75.3\eb{0.4}          & \textbf{75.6\eb{0.1}}       \\
    DD                  & 72.9\eb{1.8}          & 78.6\eb{0.4}      & 77.3\eb{0.5}  & 77.4\eb{1.2}      & 77.4\eb{1.1}      & -                     & \textbf{78.9\eb{0.1}} \\
    MUTAG               & 89.0\eb{1.1}          & 86.8\eb{1.3}      & 87.4\eb{1.0}  & 87.7\eb{0.8}      & 89.0\eb{1.3}      & 88.2\eb{1.3}          & \textbf{89.5\eb{0.1}} \\
    \bottomrule
    \end{tabular}
    
    \caption{
    Summary of the classification accuracy for transductive, inductive, heterophilic, and graph property prediction datasets, with their standard deviations. 
    These results are averaged over ten training runs with different seeds; the best results are highlighted in bold. 
    Results reproduced using GRACE with the same model as with GIRL are marked with a $^*$.
    }
    \label{tab:self_supervised_comparison}
\end{table*}

\begin{figure*}
    \centering
    \includegraphics[scale=0.12]{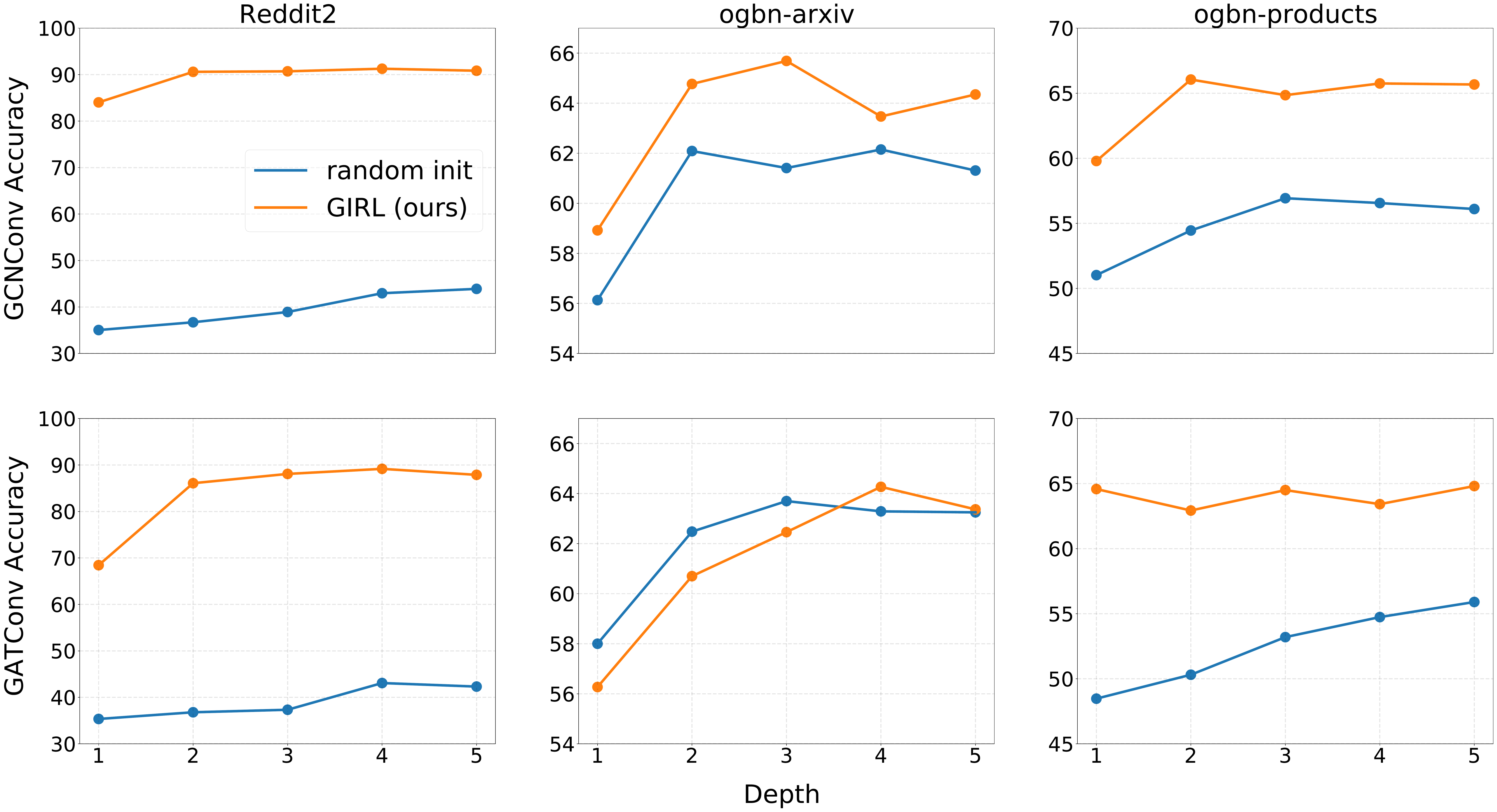}
    \vspace{-0.25cm}
    \caption{
    Comparison of downstream task test accuracy of the self-supervised setup and random initial setup.
    In both cases, the network was initialized randomly, and only in the self-supervised case, was the network trained with our self-supervised objective.
    Each column refers to a different dataset. 
    Each row corresponds to a different GNN layer type.
    }
    \vspace{-0.3cm}
    \label{fig:unsupervised_acc}
\end{figure*}

The main purpose of the following experiments is to demonstrate empirically that the KR loss minimization technique can be used for aggregation enhancement in GNNs. 
We split the experiments into two settings: supervised and self-supervised. 
In the supervised setting, we show that KR loss minimization leads to better aggregation of information, which helps to alleviate the depth problem. 
In the self-supervised setting, we compare our algorithm GIRL (\cref{fig:Self-Supervised Setting}) with other self-supervised methods.
All experiments were done on Nvidia RTX A6000 GPUs.

\subsection{Datasets}
We conducted experiments on multiple transductive (Cora, CiteSeer, Pubmed \citep{yang2016revisiting}, DBLP \citep{fu2020magnn}, Amazon-Photos \citep{shchur2018pitfalls}, WikiCS \citep{https://doi.org/10.48550/arxiv.2007.02901}, Amazon-Computers, Coauthor CS and Coauthor Physics \citep{shchur2018pitfalls}), 
inductive (Reddit \citep{hamilton2017inductive}, Reddit2 \citep{zeng2019graphsaint}, PPI \citep{zitnik2017predicting}, ogbn-arxiv, ogbn-products \citep{hu2020open}), 
heterophily (Texas, Actor \citep{https://doi.org/10.48550/arxiv.2002.05287} and USA-Airports \citep{Ribeiro_2017})
and graph property prediction datasets (NCI1, PROTEINS, DD and MUTAG \citep{https://doi.org/10.48550/arxiv.2007.08663}). 
The detailed information about the datasets can be found in \cref{tab:Datasets}. 

\subsection{Supervised Setting}
First, we study the effect of KR loss on the depth problem in GNNs, where the performance of the GNNs diminishes as they get deeper.
We show that the accuracy of deep GNNs can be improved through auxiliary KR loss, as described in \cref{eq:kr_supervised}.

\subsubsection{Experimental setup}
To demonstrate the depth problem, we trained a GNN model with different depths. The model consists of two consecutive blocks: an encoder $f_\theta$ and a decoder $p_\phi$. 
The encoder $f_\theta$ has $d$ SAGE layers \citep{hamilton2017graphsage}, where $d \in \{3, 6, 9, 12, 15, 18\}$.
In the +FA configuration, the last layer in $f_\theta$ is replaced by a fully-adjacent layer \citep{alon2021bottleneck}.
The decoder $p_\phi$ includes three fully connected layers. All but the last model layers are followed by a ReLU activation and a dropout with a drop probability of 0.1. For all datasets, we used a cluster data loader \citep{chiang2019cluster}.

\subsubsection{Results}
The results of this experiment are summarized in \cref{tab:depth_problem} and \cref{fig:depth_problem1,fig:depth_problem2}. As shown in \cref{fig:depth_problem1,fig:depth_problem2}, both training and test accuracy decrease with increasing depth, but the generalization gap does not increase. This suggests that overfitting is not a significant issue in this case. 
Instead, we observe the depth problem.
\cref{tab:depth_problem} shows that adding weight decay regularization to the supervised loss leads to a degradation in test accuracy, which is expected in the absence of overfitting. 
When, however, the KR loss is optimized alongside the supervised loss (\ref{eq:kr_supervised}), we see a significant improvement in test accuracy at all depths.
In the majority of cases, the KR loss optimization technique surpasses the +FA method \citep{alon2021bottleneck}.
We attribute the improvement in performance to the enhanced information aggregation at each GNN layer resulting from the minimization of the KR loss.

\subsection{Self-Supervised Setting}
We evaluate the effectiveness of our self-supervised graph representation learning method, GIRL, on various node classification tasks (transductive and inductive settings, as well as high-heterophily graphs) and graph classification tasks. We compare our results to those obtained by several existing self-supervised methods: DGI \citep{velivckovic2018deep}, GRACE \citep{zhu2020deep}, BGCL \citep{hasanzadeh2021bayesian}, BGRL \citep{https://doi.org/10.48550/arxiv.2102.06514} GraphCL \citep{https://doi.org/10.48550/arxiv.2010.13902}, JOAO, JOAOv2 \citep{https://doi.org/10.48550/arxiv.2106.07594}, SimGRACE \citep{Xia_2022}, SELENE \citep{zhong2022unsupervised} and GraphMAE \citep{https://doi.org/10.48550/arxiv.2205.10803}.

\subsubsection{Experimental setup}
We used a cluster data loader \citep{chiang2019cluster} to load the inductive datasets,
while the transductive and heterophily datasets 
were loaded as whole graphs. 
We used the unsupervised representation learning (URL) setting \citep{Liu_2022} and applied GIRL (\cref{fig:Self-Supervised Setting}) to a simple GNN encoder $f_\theta$ with GCN layers \citep{kipf2016semi} to obtain $\theta^*$. 
The quality of the learned encoder $f_{\theta^*}$ was then evaluated by training a decoder $p_\phi$, as described in \cref{sec:basic_notations}.

\subsubsection{Results}
The results averaged over ten training runs with different seeds are summarized in \cref{tab:self_supervised_comparison}. 
Our self-supervised method significantly outperforms existing self-supervised methods. 
GIRL can learn informative representations for both small- and large-scale graphs, while other methods often struggle with large-scale graphs. 
We believe that the inductive nature of the large graphs used in this study is the reason for the poor performance of other methods. 
It is well-known that learning on inductive graphs is more challenging than learning on transductive graphs \citep{hamilton2017inductive}. 
The existing self-supervised methods do not provide results for ogbn-products dataset. 
For comparison, we produced results (marked with a $^*$) using GRACE with the same model used in GIRL. 
We can see that our algorithm significantly improves the result on ogbn-products. 
In addition, our algorithm can be readily extended to heterophilic graphs and graph property prediction tasks as demonstrated in \cref{tab:self_supervised_comparison} (c) and (d).

\subsection{Ablation Study}
\label{ablation_study}
We compare the performance of graph representations learned using the GIRL (\cref{alg:GIRL}) to graph representations generated by a randomly initialized network in a downstream task. 
We perform this comparison using different GNN architectures and different numbers of layers. 
The encoder $f_\theta$ consists of $k \in \qty{1, 2, 3, 4, 5}$ GNN layers of one of the following types: GCNConv \citep{kipf2016semi}, GATv2Conv \citep{brody2021attentive}, GraphConv \citep{morris2021weisfeiler}, and SAGEConv \citep{hamilton2017graphsage}. 
Each GNN layer is followed by an ELU activation function. 
The decoder $p_\phi$, which is used in the downstream task, consists of two fully connected layers with ELU activation between them. 
In the self-supervised setting, the encoder $f_\theta$ is first learned using the GIRL algorithm, and then the decoder $p_\phi$ is trained on frozen features $\theta^*$, as described in \cref{sec:basic_notations}. 
In the setting with a randomly initialized network, the self-supervised training step is skipped, and the decoder $p_\phi$ is trained on frozen representations generated from randomly initialized weights $\theta$.

The results using GCNConv and GATConv layers are summarized in \cref{fig:unsupervised_acc}, with two rows corresponding to the two different layer types. The results for the remaining layer types can be found in \cref{app:rest_plots_2}. 
It is clear that graph representations learned by GIRL significantly outperform graph representations generated from a randomly initialized network. 
We also observe that in many cases, the performance of additional layers improves, demonstrating successful information aggregation from more distant nodes.

\section{Conclusion}
\label{sec:conclusion}
In this work, we present a comprehensive examination of the relationship between the KR loss and MI in the context of graph representation learning.
Through both theoretical and empirical analyses, we demonstrate that the optimization of the KR loss leads to the maximization of MI.
The convex nature of the KR method allows for efficient estimation of the KR loss without the need for additional learnable parameters, making it a more tractable approach for high-dimensional samples.
We also demonstrate the utility of the KR loss as an auxiliary loss in a supervised setting, resulting in improved representation learning at each layer of GNN. 
Our empirical results reveal that utilizing the KR loss in deeper networks effectively mitigates the depth problem and leads to improved test accuracy, even when traditional techniques such as weight decay regularization and additional fully-adjacent layers are ineffective. 
Furthermore, we introduce a novel self-supervised graph representation learning algorithm, GIRL, which is based on the KR loss. 
GIRL does not require a decoder for MI estimation or data augmentation techniques, resulting in lower complexity compared to existing self-supervised methods, and enabling unhindered scaling to larger graphs. 
Our experimental results indicate that GIRL outperforms state-of-the-art methods, particularly on large-scale graphs.

\clearpage
\bibliographystyle{plainnat}
\bibliography{references}

\clearpage
\appendix{}

\renewcommand\thefigure{\thesection.\arabic{figure}} 
\renewcommand\thetable{\thesection.\arabic{table}} 
\renewcommand\theequation{\thesection.\arabic{equation}}  
\setcounter{figure}{0}  
\setcounter{table}{0}

\crefalias{section}{appsec}
\crefalias{subsection}{appsec}
\crefalias{subsubsection}{appsec}

\section{Proofs}
\subsection{Proof of \cref{thm:t1}}
\label{proof:t1}
We prove this theorem through $\sigma$-algebras, the objects borrowed from measure theory.
A $\sigma$-algebra $\mathcal{A}$ is a collection of subsets of $\Omega$ where the following set of properties is satisfied:
\begin{enumerate}
    \item $\emptyset \in \mathcal{A}$
    \item $A \in \mathcal{A} \quad\Rightarrow\quad A^c \in \mathcal{A}$
    \item $\{ A_n \}_{n \in \mathbb{N}} \subseteq \mathcal{A} \quad\Rightarrow\quad \cup_{n \in \mathbb{N}} A_n \in \mathcal{A}$
\end{enumerate}
Each random variable $X:\Omega \rightarrow \mathbb{R}^m$ has its own $\sigma$-algebra, which is defined as follows:
\begin{align}
    \sigma_X = \{ X^{-1}(A):\; A \in \mathcal{B}(\mathbb{R}^m) \}
\end{align}
where $\mathcal{B}(\mathbb{R}^m)$ is a Borel $\sigma$-algebra on $\mathbb{R}^m$, i.e., generated from topology on $\mathbb{R}^m$.

Minimization of KR loss $\rho(Y|X) = \inf_{Z \in \mathcal{C}_X} d(Y, Z)$ by altering $X$ changes the set $\mathcal{C}_X$. 
When $\rho(Y|X)=0$, $Y$ is in $\mathcal{C}_X$ and thus continuous map $f:\mathbb{R}^m \rightarrow \mathbb{R}^n$ exists such that $f(X)=Y$. 
This leads to the following relation between $\sigma$-algebras of $X$ and $Y$:
\begin{align}
    \label{eq:eq_1}
    \sigma_Y \subseteq \sigma_X
\end{align}

On the other hand, the conventional definition of the entropy of random variable $X$ is given by expectation of a negative logarithm of the probability density function:
\begin{align}
    H(X) = \mathbb{E}[-\ln{f_X(X)}]
\end{align}
The equivalent, measure theoretic definition of entropy $H(X)$ is given by:
\begin{align}
    H(\sigma_X) = \sup_{P \subseteq \sigma_X} \sum_{A \in P} - \mathbb{P}(A)\ln{\mathbb{P}(A)}
\end{align}
where $P$ is a $\mathbb{P}$-almost partition of $\Omega$, i.e., satisfies the following:
\begin{enumerate}
    \item $\mathbb{P}(\cup_{A\in \mathbb{P}} A) = 1$
    \item $\forall A, B \in P \quad \mathbb{P}(A\cap B) = 0$
\end{enumerate}
From the above definition we can conclude that the entropy is fully dependent only on a $\sigma$-algebra of a given random variable $X$, and whenever we have the inclusion $\sigma_Y \subseteq \sigma_X$, the entropy of $Y$ is less than or equal to the entropy of $X$, i.e., $H(\sigma_X) \leq H(\sigma_Y)$.

The mutual information (MI) between two random variables $X$ and $Y$ is defined as:
\begin{align}
    \label{eq:eq_2}
    I(X;Y) = H(X) + H(Y) - H(X,Y) = \\
    = H(\sigma_X) + H(\sigma_Y) - H(\sigma (\sigma_X \cup \sigma_Y))
\end{align}
When we apply condition \cref{eq:eq_1} to \cref{eq:eq_2}, we receive:
\begin{align}
    I(X;Y)=H(Y),
\end{align}
which is the maximal MI that we can achieve by altering only random variable $X$.

\subsection{Proof of \cref{thm:t2}}
\label{proof:t2}
Let $U\subset \mathbb{R}^l$ be a compact\footnote{It is not a restrictive assumption that $U$ is a compact set since all tensor values in neural networks are bounded.} set,
\begin{align}
    C(U) = \{f:U \rightarrow \mathbb{R} \;|\; f \; \text{ is continuous function}\},
\end{align}
and
\begin{align}
    \forall \mathbf{x}_1, \mathbf{x}_2 \in U \qquad k(\mathbf{x}_1, \mathbf{x}_2) = \exp\left( -\frac{\norm{\mathbf{x}_1 - \mathbf{x}_2}^2}{2 \sigma^2} \right).
\end{align}

For each $\mathbf{x}\in U$, define continuous function $k(\mathbf{x},\cdot)=\phi_{\mathbf{x}}(\cdot)$, and construct the following functional space:
\begin{align}
    \mathcal{H}_0 = \mathrm{span}\qty(\{\phi_{\mathbf{x}}(\cdot) \;|\; \forall \mathbf{x} \in U\})
\end{align}
Define an inner product on $\mathcal{H}_0$ as follows:
\begin{align}
    \left\langle \sum_{i=1}^n a_i \phi_{\mathbf{x}_i}(\cdot), \sum_{j=1}^m b_j \phi_{\mathbf{x}_j}(\cdot) \right\rangle = \sum_{i=1}^n \sum_{j=1}^m a_i b_j k(\mathbf{x}_i, \mathbf{x}_j)
\end{align}
Let $\mathcal{H}$ be the completion of $\mathcal{H}_0$ with respect to this inner product. 
Now $\mathcal{H}$ is a reproducing kernel Hilbert space (RKHS) built from the kernel $k(\cdot, \cdot)$.

Since $k(\cdot, \cdot)$ is a universal kernel \citep{micchelli2006universal}, the set $\mathcal{H}$ is dense in $C(U)$ with respect to the supremum norm, i.e.:
\begin{align}
    \forall f \in C(U) \;\; \forall \epsilon > 0 \;\; \exists g \in \mathcal{H} \;\; \text{s.t.} \;\; \sup_{\mathbf{x}\in U}|f(\mathbf{x}) - g(\mathbf{x})| < \epsilon
\end{align}
In addition, $\mathcal{H}$ has reproducing property:
\begin{align}
    \forall f \in \mathcal{H} \;\; \forall \mathbf{x} \in U \quad f(\mathbf{x}) = \langle f, \phi_{\mathbf{x}}(\cdot) \rangle
\end{align}
and thus we have:
\begin{align}
    \rho(Y^{(i)}|X) = \inf_{Z \in \mathcal{C}_X} d(Y^{(i)}, Z) = \inf_{f \in C(U)} d(Y^{(i)}, f(X)) = \\
    = \inf_{f \in \mathcal{H}} d(Y^{(i)}, f(X)) = \inf_{f \in \mathcal{H}} d(Y^{(i)}, \langle f, \phi_X(\cdot) \rangle),
\end{align}
where $Y^{(i)}$ is the $i$-th component of random variable $Y$.

We denote the estimation of $\rho(Y|X)$ on a finite collection of samples 
$\{ \mathbf{x}_i \}_{i \in [n]}$ and $\{ \mathbf{y}_i \}_{i \in [n]}$ by $\hat{\rho}(Y|X)$
and use the distance induced from the $L_p$-norm, where $p \in [1,\infty)$. 
Thus we have:
\begin{align}
    \hat{\rho}(Y^{i}|X) = \inf_{f \in \mathcal{H}} \left( 
    \frac{1}{n}\sum_{k \in [n]}
    \left|\mathbf{y}^{(i)}_k-\langle f, \phi_{\mathbf{x}_k}(\cdot) \rangle\right|^p 
    \right)^{1/p}
\end{align}
From the \textit{representer theorem} \citep{kimeldorf1970representer}, there exists $f^* \in \mathcal{H}$ of the following form:
\begin{align}
    f^* = \sum_{i \in [n]} \alpha_i \phi_{\mathbf{x}_i}(\cdot),
\end{align}
which minimizes $\hat{\rho}(Y^{i}|X)$. Thus:
\begin{align}
    \hat{\rho}(Y^{i}|X) = \left( 
    \frac{1}{n}\sum_{k \in [n]}
    \left|\mathbf{y}^{(i)}_k-\langle f^*, \phi_{\mathbf{x}_k}(\cdot) \rangle\right|^p 
    \right)^{1/p}= \\
     = \min_{\alpha_j \in \mathbb{R}} \left( 
     \frac{1}{n} \sum_{k\in [n]}
     \left|\mathbf{y}^{(i)}_k - \sum_{j\in [n]} \alpha_j k(\mathbf{x}_j, \mathbf{x}_k)\right|^p 
     \right)^{1/p}= \\
     =\min_{\alpha \in \mathbb{R}^n} \frac{1}{n^{1/p}} \norm{\mathbf{y}^{(i)} - K \mathbf{\alpha}}_p
\end{align}
where $K_{ij} = k(\mathbf{x}_i, \mathbf{x}_j)$ is a Gram matrix and $\mathbf{y}^{(i)}$ is the vector composed of the $i$-th elements over all samples $\{ \mathbf{y}_i \}_{i \in [n]}$.

Decompose $\mathbf{y}^{(i)}$ into two parts:
\begin{align}
    \mathbf{y}^{(i)} = \mathbf{y}^{(i)}_\parallel + \mathbf{y}^{(i)}_\perp
\end{align}
where $\mathbf{y}^{(i)}_\parallel \in \Im(K)$ and $\forall \mathbf{v} \in \Im(K) \;\; \left(\mathbf{y}^{(i)}_\perp\right)^T \mathbf{v} = 0$. 
Then we have:
\begin{align}
    \hat{\rho}(Y^{i}|X) = \frac{1}{n^{1/p}} \norm{\mathbf{y}^{(i)}_\perp}_p
\end{align}
Since $K$ is positive semi-definite, we have the following eigendecomposition:
\begin{align}
    K = U \Lambda U^T
\end{align}
where columns of unitary matrix $U$ are eigenvectors of $K$ and $\Lambda$ is a diagonal matrix of eigenvalues.

Let:
\begin{align}
    \lambda_1, \lambda_2, ..., \lambda_k, 0, 0, ..., 0
\end{align}
be the descending order of eigenvalues, where $\lambda_k > 0$. Then,
\begin{align}
    \Pi = \sum_{i\in [k]} \mathbf{u}_i \mathbf{u}_i^T
\end{align}
is an orthogonal projection into $\Im(K)$ subspace. 
Consequently,
\begin{align}
    \hat{\rho}(Y^{i}|X) = \frac{1}{n^{1/p}} \norm{(\mathbb{I} - \Pi)\mathbf{y}^{(i)}}_p
\end{align}
where $\mathbb{I}$ is an identity matrix; and thus, we have:
\begin{align}
    \hat{\rho}(Y|X) = \frac{1}{m}\sum_{i\in [m]}\frac{1}{n^{1/p}} \norm{(\mathbb{I} - \Pi)\mathbf{y}^{(i)}}_p
\end{align}

\section{GIRL Algorithm}
\label{alg:GIRL}

\begin{algorithm}
	\caption{
	GIRL -- Graph Information Representation Learning
	}
	\begin{algorithmic}
    	\STATE \textbf{Input: } $\mathcal{G} = (\{\mathbf{x}_i\}_{i \in [n]}, \mathcal{E})$, $f_\theta=g_{\theta_d} \circ \cdots \circ g_{\theta_1}$
    	\STATE \textbf{Output: } $\theta^*$ 
        \FOR{batch $I \subseteq [n]$}
            \STATE $\mathbf{h}^{(0)}_i \leftarrow \mathbf{x}_i$ for each $i \in I$
            
            \FOR{layer $l \in [d]$}
                \STATE $(\{\mathbf{h}^{(l)}_i\}_{i \in I}, \mathcal{E}) 
                \leftarrow g_{\theta_l}\left( 
                (\{\mathbf{h}^{(l-1)}_i\}_{i \in I}, \mathcal{E})
                \right)$
                
                \FOR{$i \in I$}
                    \STATE sample uniformly $j \in \mathcal{N}(i)$
                    \STATE $\mathbf{z}^{(l-1)}_i \leftarrow \mathbf{h}^{(l-1)}_j$
                \ENDFOR
                
                \STATE $\hat{\rho}_l \leftarrow 
                \hat{\rho}\left(H^{(l-1)}|H^{(l)}\right) + 
                \hat{\rho}\left(Z^{(l-1)}|H^{(l)}\right)$
                
            \ENDFOR
            
            \STATE $\mathcal{L} \leftarrow \sum_{l \in [d]} \hat{\rho}_l$ 
            \STATE apply SGD step to minimize $\mathcal{L}$
        \ENDFOR
	\end{algorithmic}
\end{algorithm}

\section{Synthetic Data Experiments}
\label{exp:synthetic_data}
\begin{table}
    \centering
    \begin{tabular}{lcc} 
         \toprule
                        & $\bm{\hat{\rho}}$     & $\bm{\rho}$     \\
         \midrule
         $\rho(X|Z)$    & 0.119\eb{0.004}       & 0           \\
         $\rho(X|W)$    & 0.974\eb{0.026}       & 1           \\
         $\rho(Z|X)$    & 0.099\eb{0.013}       & 0           \\
         $\rho(W|X)$    & 0.110\eb{0.019}       & 0           \\
         \bottomrule
    \end{tabular}
    \caption{
    The estimated distance from $X$ to $\mathcal{C}_W$ is larger than the distance from $X$ to $\mathcal{C}_Z$, since there is no continuous map from $W$ to $X$.}
    \label{tab:Test_1}
\end{table}
\begin{figure}
    \centering
     \includegraphics[scale=0.25]{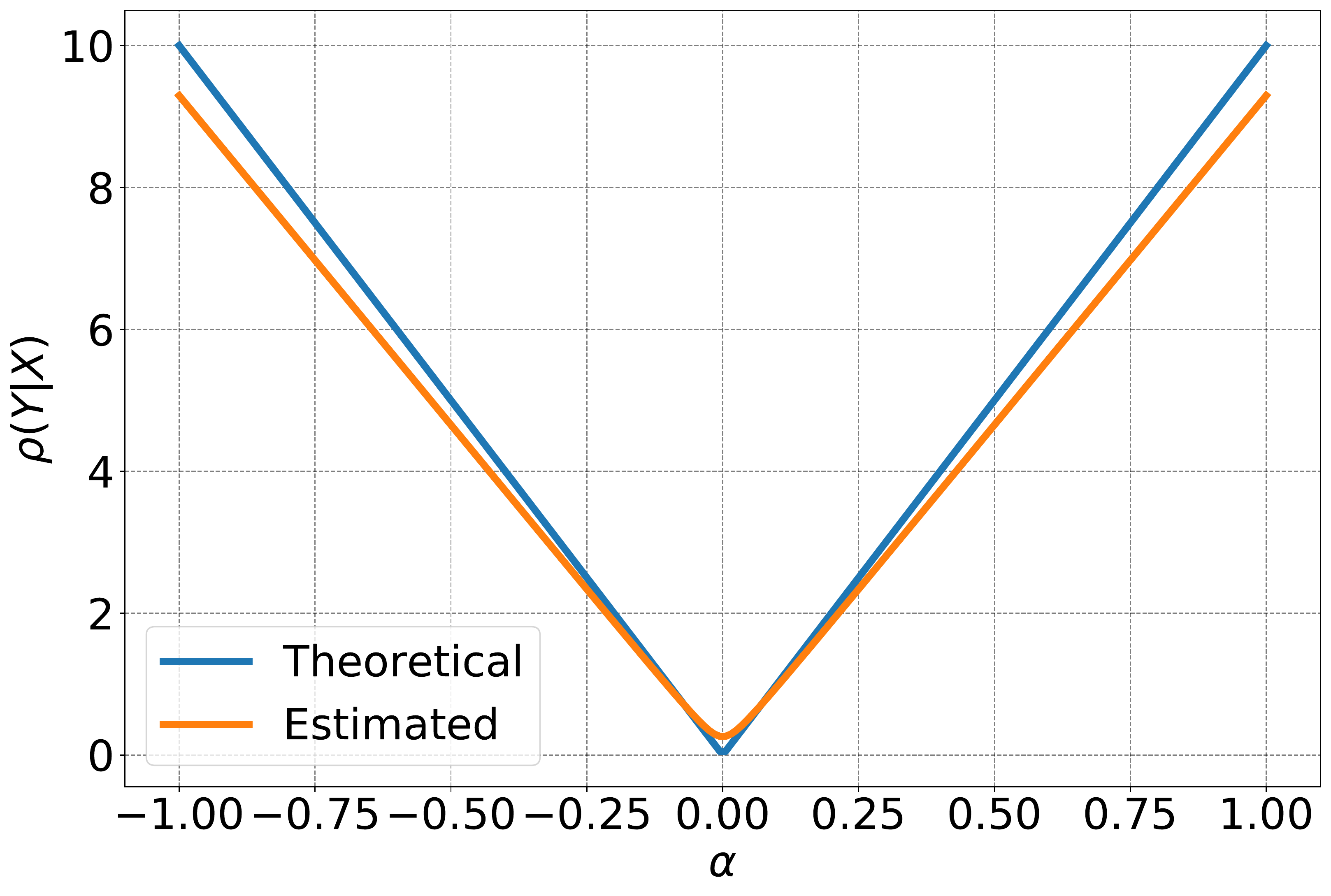}
    \caption{Comparison between the estimated $\rho(Y|X)$ and its theoretical values for different $\alpha$.}
    \label{fig:Dimensionality}
\end{figure}

\begin{figure}
    \centering
     \includegraphics[scale=0.22]{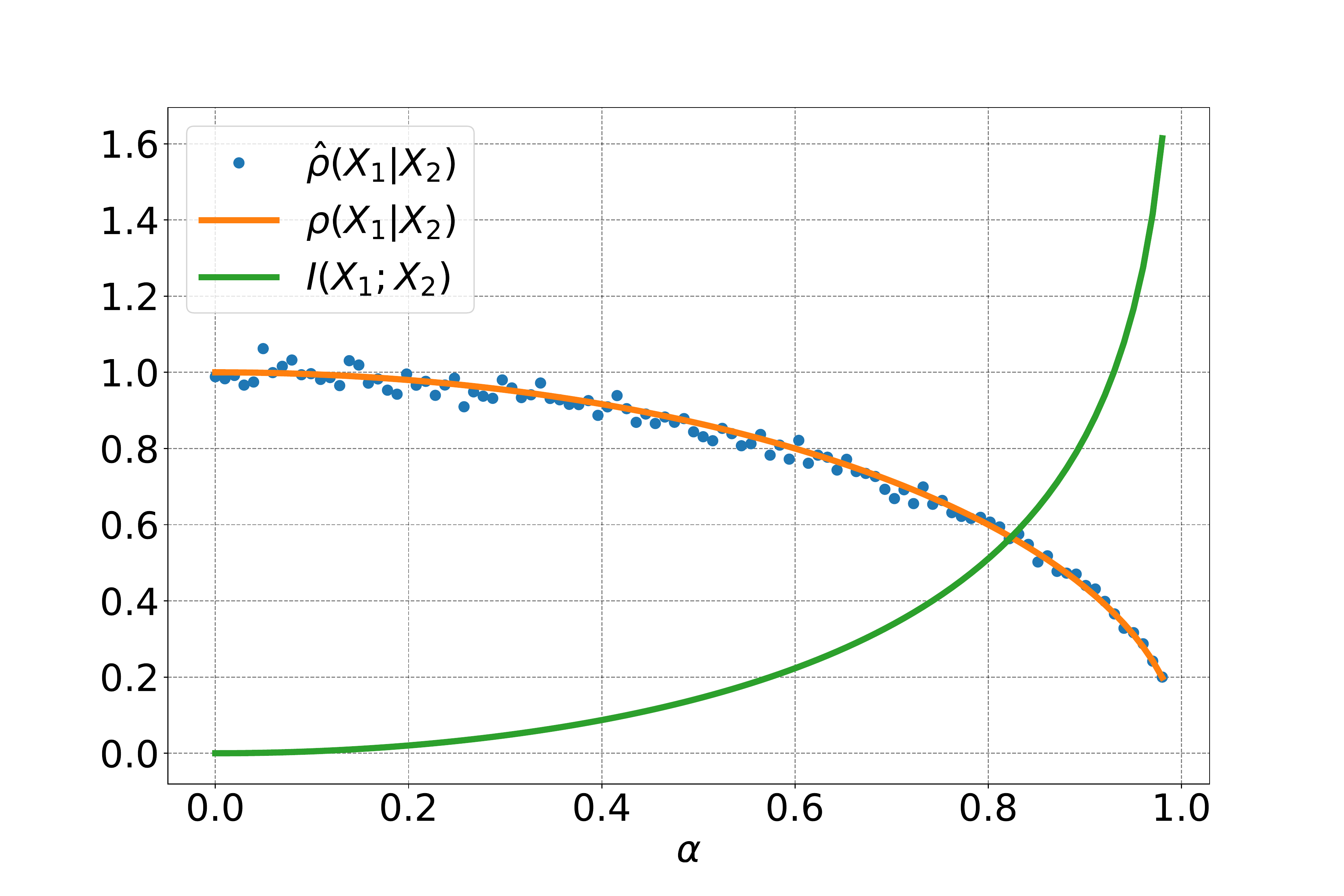}
    \caption{Comparison between the estimated $\rho(X_1|X_2)$, its theoretical values and MI, for different $\alpha$.}
    \label{fig:KL_vs_MI}
\end{figure}

If we extend the collection of continuous functions in $\mathcal{C}_X$ (\cref{eq:C_X}) to the measurable functions, and take as distance $d$ the one induced from the $L_2$ norm on random variables, then the $Z \in \mathcal{C}_X$ that minimizes $\rho(Y|X)$ is almost everywhere equivalent to conditional expectation $\mathbb{E}[Y|X]$. 
Consequently, conditional expectation $\mathbb{E}[Y|X]$ can be used to test $\rho(Y|X)$ values on synthetic data.

\paragraph{1D}
To demonstrate the ability of the KR loss to capture the existence of a continuous map from one random variable to another, we use a  simple 1D experiment. Let $X$ be a normally distributed random variable, and let $Z$ and $W$ be defined as follows:
\begin{align}
    Z = f_1(X) &= \mathrm{sign}(X) \cdot X^2\\
    W = f_2(X) &= X^2.
\end{align}
Since $f_1$ is invertible and $f_2$ is not, there exists a continuous map to $X$ from $Z$ but not from $W$. 
Of course, we there exist continuous maps to $Z$ and $W$ from $X$, since we explicitly defined continuous maps $f_1$ and $f_2$.

For this test we generated 1000 samples of $X$ and estimated $\rho$ using \cref{thm:t2}. 
The results are shown in \cref{tab:Test_1}.

\paragraph{Theoretical KR loss for \cref{tab:Test_1}}
\begin{align}
    \mathbb{E}[Z|X]=Z \quad\Rightarrow\quad \rho(Z|X) = \sqrt{\mathbb{E}[(Z-Z)^2]}=0
\end{align}
\begin{align}
    \mathbb{E}[X|Z]=X \quad\Rightarrow\quad \rho(X|Z) = \sqrt{\mathbb{E}[(X-X)^2]}=0
\end{align}
\begin{align}
    \mathbb{E}[W|X]=W \quad\Rightarrow\quad \rho(W|X) = \sqrt{\mathbb{E}[(W-W)^2]}=0
\end{align}
\begin{align}
    \mathbb{E}[X|W=w]=
\end{align}
\begin{align}
    =\mathbb{E}[\mathds{1}_{X \geq 0}X|W=w] + \mathbb{E}[\mathds{1}_{X < 0}X|W=w]=
\end{align}
\begin{align}
    =\mathbb{E}[\mathds{1}_{\sqrt{w} \geq 0}\sqrt{w}] + \mathbb{E}[\mathds{1}_{-\sqrt{w} < 0}(-\sqrt{w})]=0
\end{align}
\begin{align}
    \rho(X|W) = \sqrt{\mathbb{E}[(X-0)^2]}=1
\end{align}

\paragraph{100D}
Now, let $X$ and $N$ be 100-dimensional random vectors, with independent normally distributed entries. $Y$ be defined as
\begin{align}
    Y = \sum_{i \in [100]} (X_i + \alpha \cdot N_i),
\end{align}
where $\alpha$ is some parameter. 
At the limit, when $\alpha$ is equal to zero, there exists a continuous map from from $X$ to $Y$, and when $\abs{\alpha}$ is large, the noise $N$ dominates the value of $X$; consequently, there does not exist a continuous map from $X$ to $Y$. 
This behavior is visualized in \cref{fig:Dimensionality}, where  $\rho(Y|X)$, estimated on 1000 samples, is compared to its theoretical value.

\paragraph{Theoretical KR loss for \cref{fig:Dimensionality}}
\begin{align}
    \mathbb{E}[Y|X=\mathbf{x}] = \sum_{i \in [100]} (x_i + \alpha \cdot N_i) = \sum_{i \in [100]} x_i
\end{align}
\begin{align}
    \mathbb{E}[Y|X] = \sum_{i \in [100]} X_i
\end{align}
\begin{align}
    \rho(Y|X) = \sqrt{\mathbb{E}[(Y-\sum_{i \in [100]} X_i)^2]} = |\alpha|\sqrt{100} 
\end{align}

\paragraph{Connection between MI and KR loss}
\begin{table*}[t]
    \centering
    \begin{tabular}{lcccccc}
         \multicolumn{6}{c}{(a) \textit{Transductive}} \\
         \toprule
         \textbf{Name}                                              & \textbf{Nodes}    & \textbf{Edges}    & \textbf{Feat.\ dim.}  & \textbf{Classes}  & \textbf{Multilabel}  \\
         \midrule
         Cora \citep{yang2016revisiting}                            & 2,708             & 5,429             & 1,433                 & 7                 &--                     \\
         Citeseer \citep{yang2016revisiting}                        & 3,327             & 4,732             & 3,703                 & 6                 &--                     \\
         PubMed \citep{yang2016revisiting}                          & 19,717            & 44,324            & 500                   & 3                 &--                     \\
         DBLP \citep{fu2020magnn}                                   & 17,716            & 105,734           & 1,639                 & 4                 &--                     \\
         Amazon-Photos \citep{shchur2018pitfalls}                   & 7,650             & 119,081           & 745                   & 8                 &--                     \\
         WikiCS \citep{https://doi.org/10.48550/arxiv.2007.02901}   & 11,701            & 216,123           & 300                   & 10                &--                     \\
         Amazon-Computers \citep{shchur2018pitfalls}                & 13,752            & 245,861           & 767                   & 10                &--                     \\
         Coauthor CS \citep{shchur2018pitfalls}                     & 18,333            & 81,894            & 6,805                 & 15                &--                     \\
         Coauthor Physics \citep{shchur2018pitfalls}                & 34,493            & 247,962           & 8,415                 & 5                 &--                     \\
         \bottomrule
    \end{tabular}
    \\
     \begin{tabular}{lcccccc} 
         \multicolumn{6}{c}{} \\
         \multicolumn{6}{c}{(b) \textit{Inductive}} \\
         \toprule
         \textbf{Name}                                              & \textbf{Nodes}    & \textbf{Edges}    & \textbf{Feat.\ dim.}  & \textbf{Classes}  & \textbf{Multilabel}  \\
         \midrule
         Reddit \citep{hamilton2017inductive}                       & 232,965           & 57,307,946        & 602                   & 41                &--                     \\
         Reddit2 \citep{hamilton2017graphsage}                      & 232,965           & 11,606,919        & 602                   & 41                &--                     \\
         ogbn-arxiv \citep{hu2020open}                              & 169,343           & 1,166,243	       & 128                   & 40                &--                     \\
         PPI \citep{zitnik2017predicting}                           & 56,944            & 793,632           & 50                    & 121               &\checkmark             \\
         ogbn-products \citep{hu2020open}                           & 2,449,029         & 61,859,140        & 100                   & 47                &--                     \\
         \bottomrule
    \end{tabular}
    \\
    \begin{tabular}{lcccccc} 
         \multicolumn{6}{c}{} \\
         \multicolumn{6}{c}{(c) \textit{Heterophily}} \\
         \toprule
         \textbf{Name}                                              & \textbf{Nodes}    & \textbf{Edges}    & \textbf{Feat.\ dim.}  & \textbf{Classes}  & \textbf{Multilabel}  \\
         \midrule
         Texas \citep{https://doi.org/10.48550/arxiv.2002.05287}    & 183               & 325               & 1,703                 & 5                 &--                     \\
         Actor \citep{https://doi.org/10.48550/arxiv.2002.05287}    & 7,600             & 30,019            & 932                   & 5                 &--                     \\
         USA-Airports \citep{Ribeiro_2017}                          & 1,190             & 13,599	           & 1,190                 & 4                 &--                     \\
         \bottomrule
    \end{tabular}
    \\
    \begin{tabular}{lcccccc}
         \multicolumn{6}{c}{} \\
         \multicolumn{6}{c}{(d) \textit{Graph Property Prediction}} \\
         \toprule
         \textbf{Name}                                              & \textbf{Graphs}   & \textbf{Avg. Nodes}   & \textbf{Avg. Edges}   & \textbf{Feat.\ dim.}  & \textbf{Classes}  \\
         \midrule
         NCI1 \citep{https://doi.org/10.48550/arxiv.2007.08663}     & 4110              & 29.87                 & 32.30                 & 37                    & 2                 \\
         PROTEINS \citep{https://doi.org/10.48550/arxiv.2007.08663} & 1113              & 39.06                 & 72.82                 & 3                     & 2                 \\
         DD \citep{https://doi.org/10.48550/arxiv.2007.08663}       & 1178              & 284.32                & 715.66                & 89                    & 2                 \\
         MUTAG \citep{https://doi.org/10.48550/arxiv.2007.08663}    & 188               & 17.93                 & 19.79                 & 7                     & 2                 \\
         \bottomrule
    \end{tabular}
    \caption{Node property prediction dataset statistics.}
    \label{tab:Datasets}
\end{table*}
The following experiment demonstrates the relationship between MI between random variables and their KR loss.
Let $X_1$ and $X_2$ be two normally distributed random variables with $Cov(X_1, X_2) = \alpha$.
In this case, the MI between $X_1$ and $X_2$ can be calculated theoretically and given by: 
\begin{align}
    I(X_1; X_2) = - \frac{1}{2} log(1 - \alpha^2)
\end{align}
The theoretical KR loss is given by:
\begin{align}
    \rho(X_1|X_2) = \sqrt{1 - \alpha^2}
\end{align}
In \cref{fig:KL_vs_MI}, we presented a comparison between the theoretical MI, KR loss and estimated KR loss.
First, we see again that the theoretical KR loss overlaps with the estimated loss. 
Second, as $\alpha$ grows the correlation between $X_1$ and $X_2$ is stronger and the KR loss tends to zero whereas MI increases.

\section{Dataset Statistics}
The datasets used in our experiments are given in \cref{tab:Datasets}.

\section{Additional Experimental Results}
\label{app:rest_plots_2}
\begin{figure*}
    \centering
    \includegraphics[scale=0.12]{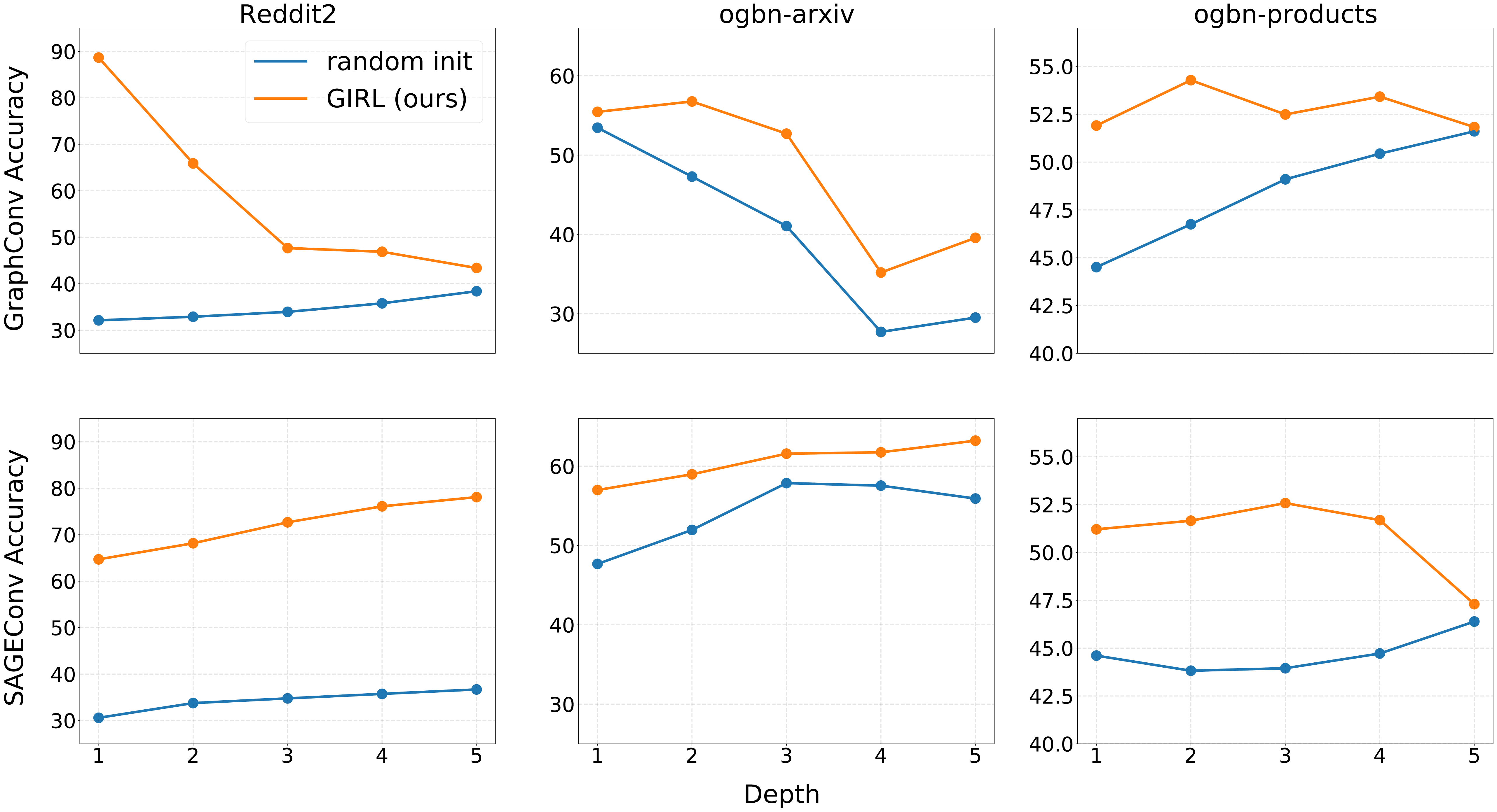}
    \caption{
    Comparison of test accuracy in the downstream tasks of the self-supervised setup and random initial setup, where in the random initial setup, the self-supervised training was skipped. 
    Each column refers to a different dataset. 
    Each row corresponds to a different GNN layer type.
    }
    \label{fig:rest_unsupervised_acc}
\end{figure*}

In \cref{fig:rest_unsupervised_acc} we demonstrate the results for the GraphConv and SAGEConv layers, which are related to the ablation study in \cref{ablation_study}.

\end{document}